%% file: root.tex
\documentclass[letterpaper, 10 pt, conference]{ieeeconf}  

\IEEEoverridecommandlockouts                              

\usepackage{cite}

\usepackage{amsmath,amsfonts}
\usepackage{algorithmic}
\usepackage{array}
\usepackage[caption=false,font=normalsize,labelfont=sf,textfont=sf]{subfig}
\usepackage{textcomp}
\usepackage{stfloats}
\usepackage{url}
\usepackage{verbatim}
\usepackage{graphicx}
\usepackage{balance}
\usepackage{booktabs}
\usepackage{multirow}
\usepackage{xcolor}
\usepackage{cleveref}
\usepackage{bm}
\usepackage{amsmath}
\usepackage{siunitx}
\usepackage{algorithm}

\usepackage[switch, columnwise]{lineno}

\title{\LARGE \bf
CommCP: Efficient Multi-Agent Coordination via LLM-Based Communication with Conformal Prediction
}

\author{Xiaopan Zhang$^{*}$, Zejin Wang$^{*}$, Zhixu Li, Jianpeng Yao, Jiachen Li$^{\ddag}$
\thanks{$^*$Equal contribution \ $^\ddag$Corresponding author}
\thanks{X. Zhang, Z. Wang, Z. Li, J. Yao, and J. Li are with the Trustworthy Autonomous Systems Laboratory at the University of California, Riverside, CA, USA. {\tt\small \{xzhan006, jiachen.li\}@ucr.edu}.}
}

\begin{document}
\linespread{0.97}
\maketitle
\thispagestyle{empty}
\pagestyle{empty}

\begin{abstract}
    \input{contents/00abstract}
\end{abstract}

\section{Introduction}\label{sec:introduction}
    \input{contents/01intro}
    
\section{Related Work}\label{sec:relatedwork}
    \input{contents/02relatedwork}

\section{Problem Formulation}\label{sec:problemformulations}
    \input{contents/03problemformulations}

\section{Method}\label{sec:multi_task_navigation}
    \input{contents/04methods}

\section{Experiments}\label{sec:experiments}
    \input{contents/05experiments}

\section{Conclusion}\label{sec:conclusions}
    \input{contents/06conclusions}

\bibliographystyle{IEEEtran}
\bibliography{References}
\end{document}

%% file: contents/00abstract.tex
To complete assignments provided by humans in natural language, robots must interpret commands, generate and answer relevant questions for scene understanding, and manipulate target objects.
Real-world deployments often require multiple heterogeneous robots with different manipulation capabilities to handle different assignments cooperatively. 
Beyond the need for specialized manipulation skills, effective information gathering is important in completing these assignments. 
To address this component of the problem, we formalize the information-gathering process in a fully cooperative setting as an underexplored multi-agent multi-task Embodied Question Answering (MM-EQA) problem, which is a novel extension of canonical Embodied Question Answering (EQA), where effective communication is crucial for coordinating efforts without redundancy.
To address this problem, we propose \textbf{CommCP}, a novel LLM-based decentralized communication framework designed for MM-EQA. 
Our framework employs conformal prediction to calibrate the generated messages, thereby minimizing receiver distractions and enhancing communication reliability.
To evaluate our framework, we introduce an MM-EQA benchmark featuring diverse, photo-realistic household scenarios with embodied questions. 
Experimental results demonstrate that \textbf{CommCP} significantly enhances the task success rate and exploration efficiency over baselines. The experiment videos, code, and dataset are available on our project website: \url{https://comm-cp.github.io}.

%% file: contents/01intro.tex
Modern service robots are designed to understand human instructions and complete tasks in real-world household environments (e.g., ``Turn off the TV if it is currently on," ``Bring the red pillow from the living room to the bedroom"). This process involves interpreting natural language commands, generating and answering relevant questions for scene understanding and reasoning (e.g., ``Is the TV turned on?" ``What is the color of the pillow?"), and manipulating target objects accordingly. 
A crucial step is answering these questions, a task known as Embodied Question Answering (EQA)~\cite{das2018embodied}, which requires robots to efficiently explore the 3D environment from a random starting location and actively gather information until a confident answer can be provided. 
Prior studies have investigated this in single-agent settings \cite{das2018embodied,ren2024explore,majumdar2024openeqa,yu2019multi,yao2025towards}. In contrast, we envision future households with multiple \textit{heterogeneous} service robots, each with distinct capabilities and non-transferable assignments. While they cannot take over each other's tasks, they can access all generated questions and share observations and interpretations to enhance exploration efficiency. We define this cooperative information-gathering setting as a \emph{multi-agent multi-task EQA (MM-EQA)} problem, a novel challenge that facilitates multi-robot collaboration in real-world scenarios.

\begin{figure}[!tbp]
    \centering
    \includegraphics[width=1\linewidth]{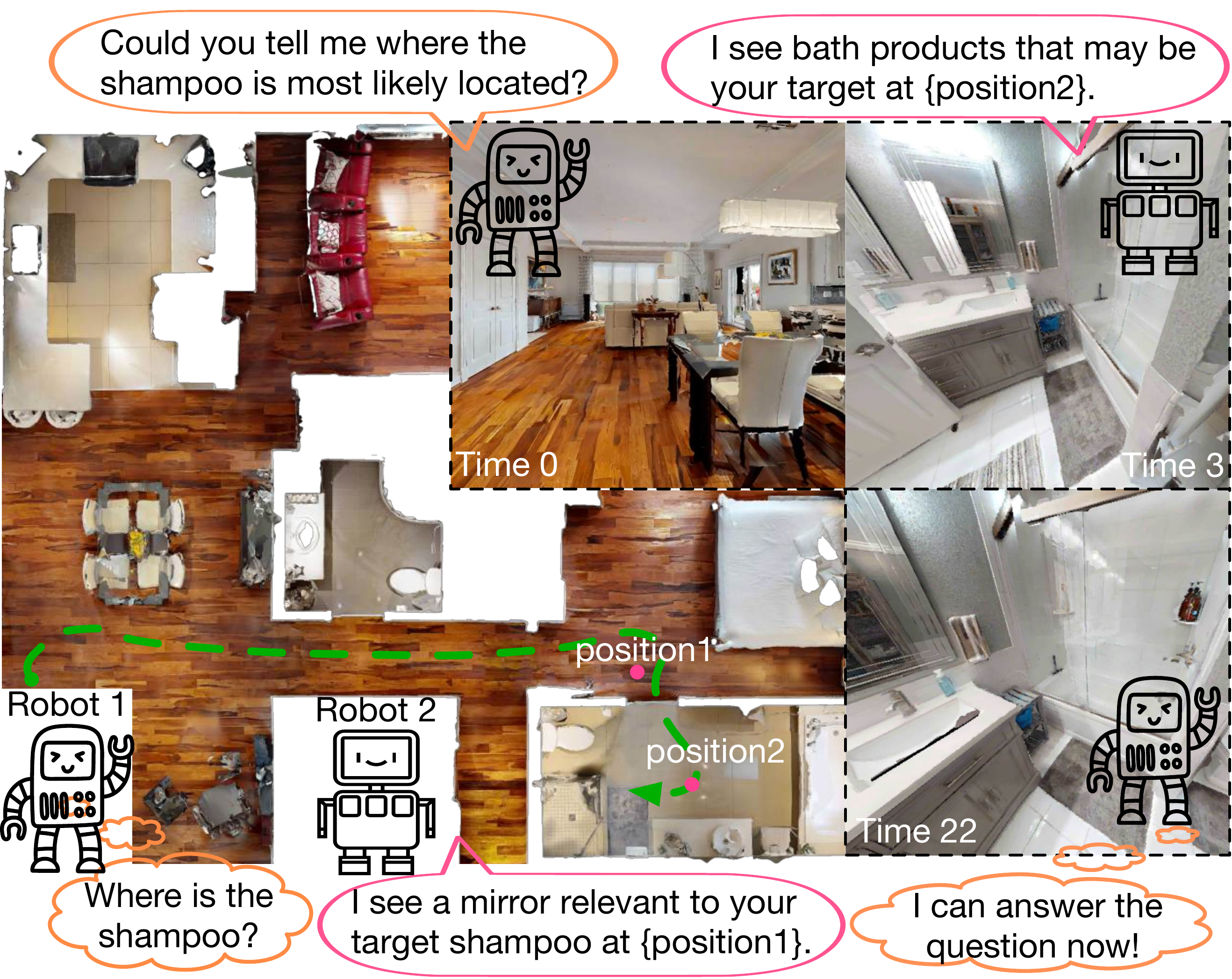}
    \vspace{-0.7cm}
    \caption{In a household setting, robots exchange observations and reasoning to collaboratively complete their assigned tasks. Each agent generates confident and goal-directed messages using calibrated outputs from LLMs. The bottom-left image shows a bird’s-eye view of Robot 1’s navigation path after incorporating information received from Robot 2. The top-right sequence captures both robots’ camera views at different timestamps.}
    \vspace{-0.7cm}
    \label{fig:intro}
\end{figure}

While existing single-agent EQA solutions can be adapted to a multi-agent setting by having each robot work independently, this naive approach is inefficient. 
Communication enables mutual assistance, improving exploration and increasing the likelihood of faster task completion. 
However, uncalibrated communication could hinder efficiency by sharing irrelevant or misleading information. 
Therefore, it is critical to ensure that messages are accurate and pertinent to the recipient's tasks.
This work tackles the MM-EQA problem by designing a communication framework that enhances multi-agent exploration efficiency and task performance.

Large language Models (LLMs) have shown great potential in solving EQA tasks due to their remarkable ability to understand natural-language queries, reason, and provide answers in natural language~\cite{zhang2023building}. 
In the context of MM-EQA, natural language is an ideal communication protocol, as LLMs are inherently trained to engage in dialogues. Several LLM-based communication methods have been proposed in other domains~\cite{zhang2024towards,rasal2024llm}, but these cannot be directly adapted to our MM-EQA setting. 
Additionally, LLMs often produce miscalibrated and overconfident outputs~\cite{ mielke2022reducingconversationalagentsoverconfidence}, which can result in irrelevant or misleading information. 
This can hinder cooperation efficiency, as agents may share inaccurate data, reducing overall exploration effectiveness~\cite{hao2024quantifying}.

Our work tackles this challenge and develops an LLM-based communication framework for MM-EQA. Our key insight is that \emph{an agent should only communicate information it confidently deems relevant to its partner agents' tasks} (see Fig. \ref{fig:intro}).
We propose CommCP, a novel decentralized LLM-based communication framework that employs conformal prediction (CP) \cite{shafer2008tutorial,ren2023robots} to calibrate the confidence of LLM's outputs. 
Our framework ensures that the outputs generated by LLMs are more reliable and reduces the negative impact of irrelevant or misleading information to partner agents, ultimately enhancing the overall task performance and efficiency of the multi-agent system.
To evaluate our proposed framework, we create a novel MM-EQA benchmark based on realistic scenarios and the Habitat-Matterport 3D (HM3D) dataset \cite{ramakrishnan2021habitat}.
The experimental results show that our approach enhances the task success rate and shortens completion time by a large margin over baselines.

The main contributions of this paper are as follows:
\begin{itemize}
    \item We formulate the information-gathering process of completing assignments provided in natural language as a novel multi-agent multi-task embodied question answering (MM-EQA) problem, where multiple robots work as a team, handling EQA tasks in a shared environment and communicating to exchange information or answers.
    \item We propose CommCP, a novel LLM-based decentralized communication framework for MM-EQA, where conformal prediction is employed to calibrate the generated messages to reduce distractions to other agents and improve communication reliability and efficiency. 
    \item We create a novel MM-EQA benchmark with photo-realistic scenarios from the HM3D dataset to validate the effectiveness of the proposed framework. This benchmark is released to facilitate future studies.
\end{itemize}

%% file: contents/02relatedwork.tex
\subsection{LLM-based Decentralized Multi-Agent Cooperation}
LLM-based multi-agent cooperation has gained increasing attention recently \cite{chan2023chateval,zhang2024lamma}, with various systems developed for multi-agent tasks \cite{hong2023metagpt,horton2023large,hua2023war,zhang2024towards,chen2024scalable,mandi2024roco}. 
Unlike single-agent or centralized systems, decentralized cooperative systems involve peer-to-peer communication, where agents interact directly, an architecture common in world simulation applications \cite{zhang2024agentcf,zhou2023sotopia}.
In these systems, communication typically takes the form of natural language text generated by LLMs, with content varying by application, such as sharing environmental observations, coordinating actions, or reallocating tasks. 
However, the effectiveness of LLM-generated communication remains underexplored. As noted in \cite{zhang2023building}, decentralized communication often incurs costs such as bandwidth limitations or delays. Thus, agents must communicate efficiently and avoid unnecessary or redundant messages. Current approaches lack mechanisms to assess communication quality and rely solely on raw LLM outputs, leading to inefficiencies, especially when agents act on incomplete or uncertain information.
\subsection{Conformal Prediction and Calibration}
Recent research has highlighted the miscalibration issue in LLMs, where models may exhibit overconfidence or under-confidence in their text outputs. 
This presents a huge challenge as foundation models are applied to embodied tasks where agents may have miscalibrated confidence in their decisions. Previous work \cite{ren2023robots,wang2024safe} has employed conformal prediction  \cite{shafer2008tutorial} to formally quantify an LLM’s uncertainty in a robot planning context, which ensures that the robot’s plans are executed with calibrated confidence.
Explore until Confident~\cite{ren2024explore} extends this approach by applying multi-step conformal prediction in EQA tasks to determine when the VLM is sufficiently confident when a visual language model (VLM) is sufficiently confident to stop exploration.
To our best knowledge, we are the first to employ conformal prediction to enhance multi-agent communication through calibrating confidence during collaborative exploration, which is a setting not addressed by prior work.

%% file: contents/03problemformulations.tex
Consider a scenario where $N_a$ robots are deployed in a 3D scene with multiple different assignments, each starting from an initial pose $g^i_0$, and aiming to answer the questions $q^i_{1:N_q}$ related to its assignments.
The objective is to maximize the success rate while minimizing the exploration time, with all answers required within a time horizon $T_\text{max}$.
Each robot knows all questions, including those assigned to others.
They can communicate via natural language messages, denoted $\zeta^i$, to exchange information.

Each robot $i\in N_a$ is equipped with cameras that, at each time step $t$, can provide the robot with an RGB image $I^i_{c,t}$ and a depth image $I^i_{d,t}$ of the local scene as observations. The pose (2D position and orientation) of each robot at each time step is denoted as $g^i_t$, with the poses of all robots collected into a set $G_t = \{g^i_t \mid i=1,...,N_a\}$. Each robot is equipped with a collision-free planner $\pi$ to navigate. Given the current pose $g^i_{t}$ and a target position, the planner $\pi$ determines the next feasible pose $g^i_{t+1}$, with a low-level controller transporting the robot to the planned pose at $t+1$. 

In this case, a multi-robot multi-task Embodied Question Answering (MM-EQA) problem is defined with a tuple $\xi:= (E, G_0, T_\text{max}, Q, Y)$, where $E$ is the 3D scene with dimensions $L \times W \times H$, which is discretized into a voxel map $M$ composed of cubes with a side length of $l$. $L$, $W$, and $H$ representing the length, width, and height of the voxel map $M$; $G_0 = \{g^i_0 \mid i=1,...,N_a\}$ is a set of initial poses of the robots, and $T_\text{max}$ is the maximum time horizon allowed for the robots to explore the scene and complete the task. Each robot $i$ is assigned with $N_q$ questions to answer. The set $Q = \{q^i_j \mid i=1,...,N_\text{a},j=1,...,N_\text{q}\}$ collects all the questions assigned to the robots, with $q^i_j$ being the $j^\mathrm{th}$ question assigned to the $i^\mathrm{th}$ robot. Each question is a multiple-choice question with four choices $\{\text{`A', `B', `C', `D'}\}$. The ground truth answers are denoted by the set $Y = \{a^i_j\in \{\text{`A', `B', `C', `D'}\} \mid i=1,...,N_a,j=1,...,N_q\}$.

%% file: contents/04methods.tex
\begin{figure*}[!tbp]
	\centering
        \includegraphics[width=\linewidth]{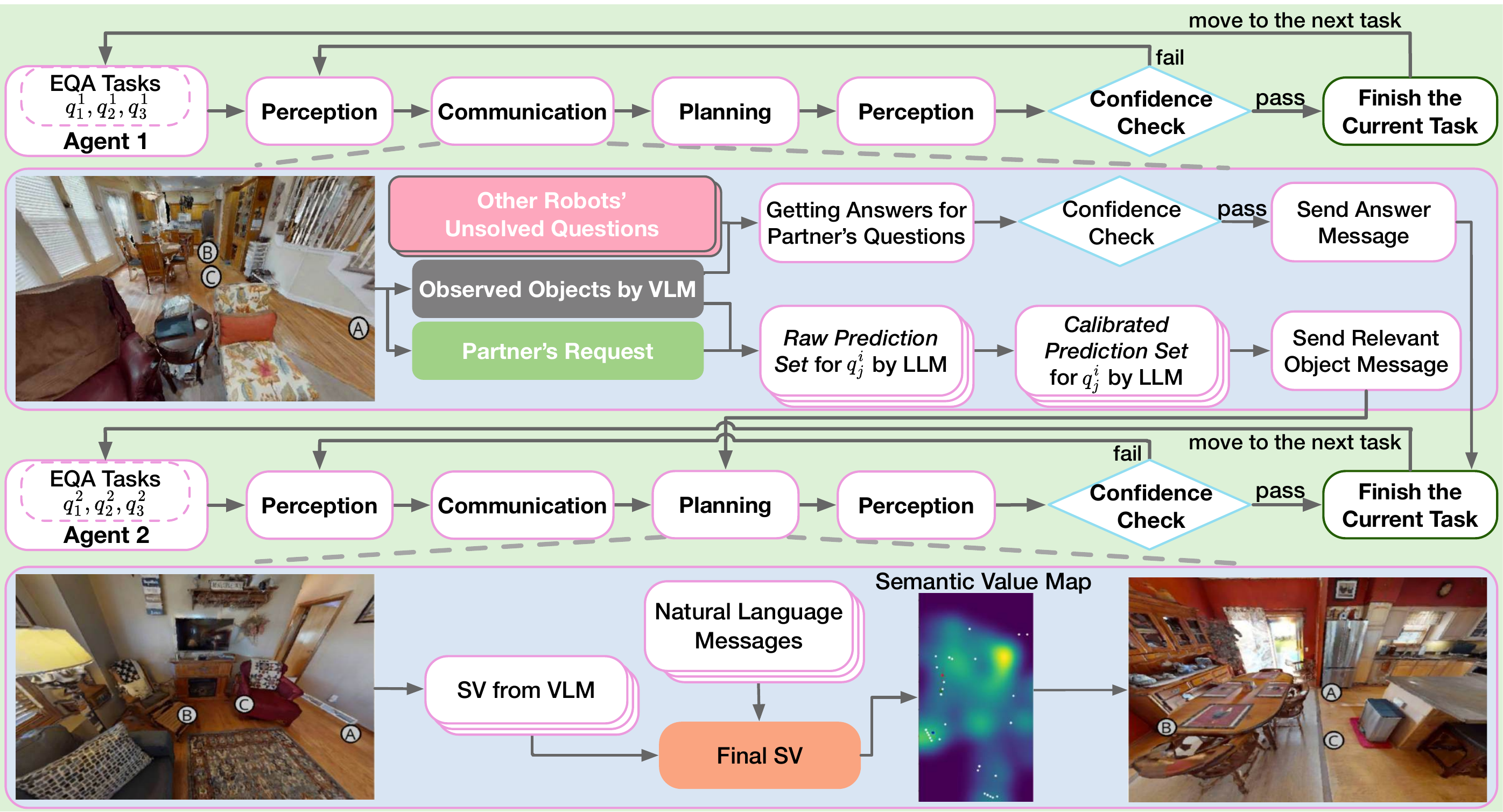}
        \vspace{-0.5cm}
	\caption{An overview of our framework shows each robot with a perception module, a communication module, a planning module, and a confidence check module. At each time step, a robot generates local and global semantic values (SV) based on the current view and the communication message from the other robot. It navigates using a 2D weighted semantic value map and handles related object-check requests from other agents. The messages are generated based on the robot's current view, which are calibrated by conformal prediction to enhance relevance.}
    \vspace{-0.3cm}
    \label{fig:process}
\end{figure*}

This section introduces the CommCP framework, which leverages communication to enhance multi-agent exploration for the MM-EQA problem.
Building upon the approach presented in \cite{ren2024explore} for single-agent single-task EQA, our framework further enables communication capabilities to improve task completion efficiency and success rate. Furthermore, our framework can be easily extended to handle more complex human assignments in multi-agent settings, where robots are tasked with executing downstream tasks based on the information they acquire since robots are assigned questions based on their capabilities for the downstream tasks.

The overall framework architecture is illustrated in Fig.~\ref{fig:process}, which consists of four core modules: perception, communication, planning, and confidence check modules. The planning and confidence check modules are modified from \cite{ren2024explore}. In the communication module, messages are generated and projected onto the semantic value map in the planning module to guide the robots' exploration strategies. This module also enables a robot to provide answers to other robots' questions when it has sufficient confidence. All notations in this section correspond to time $t$.
For each robot $i$, the perception module employs a VLM to detect a set of observed objects $O^{i}_{\text{observe}}$ from the RGB image $I^i_{c}$ with the following prompt:

\vspace{+0.1 cm}
\colorbox{lime!20}{
\parbox{\dimexpr0.90\linewidth-2\fboxsep}{
\fontsize{8.5pt}{\baselineskip}\selectfont
\textit{
Consider the indoor scenario, analyze the provided image and list all the objects you observe. Provide the name of each object along with its color, separated by a comma. 
}
}}
\vspace{+0.1 cm}

The perception module's detected $O^{i}_{\text{observe}}$ are fed into the reasoning process to determine which objects to include in the communication by employing conformal prediction.
Each unsolved question is sequentially prompted to the LLM, along with 
$O^{i}_{\text{observe}}$, to generate an answer. 
If the answer passes the confidence check described in Sec. IV.D and the question is assigned to the responding robot, it proceeds. Otherwise, it sends the answer to the responsible robot. 
\subsection{LLM-Based Object Relevance Reasoning}\label{sec:A}
In indoor scenarios, objects are typically organized according to patterns of human usage, and LLMs can leverage the general knowledge of these patterns. 
Thus, if LLM assesses that an object observed in an area is highly relevant to the target object, there is an increased likelihood that the target is nearby. Based on this intuition, we design a communication framework that enables robots to share relevant object information or answers to other robots' questions.

During exploration, each robot $\hat{i}$ sends a request $r^{\hat{i}}_j$ to seek for assistance on its question $q^{\hat{i}}_j$ and provide its target objects $O^{\hat{i}}_{\text{request}}$. 
The LLM evaluates objects observed by robot $i$, denoted as $O^{i}_{\text{observe}}$, and the observed and target objects are labeled as \textbf{Observed} and \textbf{Request}, respectively. We employ the zero-shot chain-of-thought \cite{kojima2022large} to prompt the LLM to conduct detailed reasoning before generating a final output. 
The following is the prompts used for the LLM. Here, the \textit{system prompt} is the instruction for the LLM, and the \textit{user prompt} is the content of the conversation with LLM.

\vspace{+0.1 cm}
\colorbox{lime!20}{
\parbox{\dimexpr0.90\linewidth-2\fboxsep}{
\raggedright
\fontsize{8.5pt}{\baselineskip}\selectfont
\textbf{System prompt:} \\
\textit{As a robot in a house, your partner is looking for \{\textbf{Request}\} and you can inform them about what you have observed.}
}}

\colorbox{lime!20}{
\parbox{\dimexpr0.90\linewidth-2\fboxsep}{
\fontsize{8.5pt}{\baselineskip}\selectfont
\textbf{User prompt:} \\
\textit{You observe \{\textbf{Observed}\}. Your partner wants to find \{\textbf{Request}\}. Since you and your partner are not in different locations, evaluate whether it is worth it for your partner to travel to your position to find \{\textbf{Request}\}.\\ 
You have four options: \textbf{A} (Yes. Because \{\textbf{Request}\} and \{\textbf{Observed}\} are same, 
and you might directly find \{\textbf{Request}\}).}
\textit{
\textbf{B} (Yes. They are highly relevant. This means \{\textbf{Request}\} should be close to \{\textbf{Observed}\}).
\textbf{C} (No. They are not strongly related).   
\textbf{D} (No. \{\textbf{Observed}\} is a common feature).}

\textit{For option A, consider if \{\textbf{Observed}\} and \{\textbf{Request}\} are the same things.  
For option B, consider where \{\textbf{Request}\} is most likely to be found. Then evaluate if \{\textbf{Observed}\} is likely to appear in that location as well. 
For options C and D, since you and your partner are not in the same position, assess if \{\textbf{Observed}\} is worth passing by. 
Option D implies that \{\textbf{Observed}\} is just a common feature in the house. 
Now provide your analysis.}
}}
\vspace{+0.13 cm}

The LLM's output, \textit{\textbf{Analysis}}, is used to compute probabilities for each option. We employ the following prompt that directs the LLM to select only one option based on \textit{\textbf{Analysis}}.

\vspace{0.13cm}
\colorbox{lime!20}{
\parbox{\dimexpr0.90\linewidth-2\fboxsep}{
\fontsize{8.5pt}{\baselineskip}\selectfont
\textbf{System prompt:} \\
\textit{You should only output one letter.} \\
\textbf{User prompt:} \\
\textit{
Your observe $O^{\text{observe},i}_{t,k}$ now. Your partner wants to find \{\textbf{Request}\}. 
Since you are in different locations, you need to assess if it's worthwhile for your partner to come for \{\textbf{Request}\}. 
You have four options: \textbf{A, B, C, D}. Here is your previous analysis: \{\textbf{Analysis}\}. You should only output one letter.}
}}
\vspace{0.13cm}

Each observed object for robot $i$ is assigned an \textit{option-probability pair} $O^{i}_{\text{observe},k} := \{\text{Option}_k, p_k \mid k=1,...,N_{k}\}$, where $N_k$ is the number of observed objects and $p_k$ is the probability of the token corresponding to $\text{Option}_k$ generated by the LLM. 
If  \textbf{Option A} is returned, it is likely to correspond to the target object $O^{i}_{\text{target},j}$ of request $r^{\hat{i}}_j$. 
If \textbf{Option B} is selected, it is likely a relevant object $O^{i}_{\text{relevant},j}$ related to request $r^{\hat{i}}_j$. 
Objects with options C and D are disregarded.

\subsection{Message Generation with Conformal Prediction}

To mitigate potential miscalibration in LLM-generated lists of targets and relevant objects, which leads to irrelevant information, we employ conformal prediction (CP) \cite{shafer2008tutorial} to calibrate object lists for message generation. 
CP provides statistically guaranteed prediction sets that contain the ground truth with a user-specified probability \cite{ren2023robots}.
Specifically, we adopt split conformal prediction \cite{fontana2023conformal}, which involves designing a conformity score to measure the reliability of predictions, collecting a calibration set to establish standards for comparing conformity scores during testing, and determining whether to include a prediction result in the final prediction set. 
In the context of deep learning, the conformity score is defined as the probability output of models as this reflects its confidence in the answers it produces.

In our framework, we adopt $p_k$ as conformity scores, and only $\text{Option}_k$ with $p_k$ higher than a computed threshold are included in the calibrated prediction set. To handle different probability distributions for \textbf{Options A} and \textbf{Options B}, we define two calibration sets: \( \mathcal{Z}^{A}_{\text{cal}} = \{ z_k = (\text{`A'}, p_k) \mid k=1,..., N_k\} \) and \( \mathcal{Z}^{B}_{\text{cal}} = \{ z_k = (\text{`B'}, p_k) \mid k=1,..., N_k\} \). 
To collect the calibration sets, we sample (observed\_object, target\_object) pairs from 20 diverse HM3D scenarios and generate ground truth labels. To handle scenarios where multiple objects may be relevant, we formally define the ground truth for calibration as the single option with the highest LLM confidence score, following the procedure in \cite{ren2023robots}. For each labeled pair, we apply the LLM reasoning process described in Section V.A to obtain probability values $p_A$ and $p_B$, which form the calibration sets $\mathcal{Z}^{A}_{\text{cal}}$ and $\mathcal{Z}^{B}_{\text{cal}}$, respectively. 
These calibration datasets represent samples from the underlying distributions $\mathcal{D}^{A}_{\text{cal}}$ and $\mathcal{D}^{B}_{\text{cal}}$ for spatial co-occurrence judgments in household environments. 
Let $\mathcal{D}^{A}_{\text{test}}$ and $\mathcal{D}^{B}_{\text{test}}$ represent the unknown distributions of option-probability pairs for a new scenario, which are assumed to be exchangeable with $\mathcal{D}^{A}_{\text{cal}}$ and $\mathcal{D}^{B}_{\text{cal}}$.

In our MM-EQA setting, we argue that calibration and test samples can be treated as independent and identically distributed (i.i.d). Let $(s_i, p_i)$ represent the $i$-th sample, where $s_i$ denotes the spatial relationship (A or B) and $p_i$ denotes the LLM's probability output. Our i.i.d. assumption states:
\vspace{-0.15cm}
\begin{equation}
P((s_1, p_1), (s_2, p_2), \ldots, (s_n, p_n)) = \prod_{i=1}^{n} P(s_i, p_i),
\vspace{-0.15cm}
\end{equation}
where $(s_i, p_i) \sim F$ for all $i$, and $F$ is a fixed joint distribution over spatial relationships and probability values. 
This assumption holds because the calibration process satisfies both independence and identical distribution requirements: 
1) Calibration scenarios are randomly sampled from the HM3D household dataset without temporal or spatial dependencies; 
2) Each (observed\_object, target\_object) evaluation represents an independent semantic judgment; 
3) LLM's probability outputs for different object pairs are not causally related; 
4) Spatial co-occurrence patterns represent typical household organization principles; 
5) The LLM's semantic understanding remains consistent across scenes; 
6) The underlying distribution $F$ of object spatial relationships is preserved across the dataset; 
and 7) During testing, each robot generates semantic values for target and relevant objects based solely on its current view, without considering past frames, which ensures spatial and temporal independence required for conformal prediction.
Fulfilling these conditions ensures that the prerequisite for applying conformal prediction is met.
While CP typically requires only exchangeability of data, our i.i.d. assumption provides stronger theoretical guarantees. 
Approximately $1/\delta$ calibration samples are needed for reliable coverage~\cite{luo2024sample}, and 20 scenarios provide adequate data to achieve typical confidence levels (e.g., $\delta = 0.05$).

For a new test sample \( z_{\text{test}} = (\text{`A' or `B'}, p_{\text{test}}) \), we use CP to construct a prediction set $\mathcal{C}(z_{\text{test}})$ using a calibrated score threshold, $p_{\text{thres}}$. 
This threshold is set to the $1-\epsilon_1$ quantile of the scores in the calibration set, where \( \epsilon_1 \) is a user-defined miscoverage rate (e.g., $\epsilon_1=0.05$ for 95\% confidence) that influences the size of the prediction set \( \mathcal{C}(\cdot) \). 
The prediction set is then formed by including all options whose scores, $p_{\text{test}}$, meet or exceed this threshold. 
This construction provides the formal marginal guarantee that the true label is contained within the prediction set with high probability:
\vspace{-0.15cm}
\begin{equation}
    P(z_{\text{test}} \in \mathcal{C}(z_{\text{test}})) \geq p_{\text{thres}}
    \vspace{-0.15cm}
\end{equation}
A higher $p_{\text{thres}}$ leads to more confident options being shared during communications, and vice versa.
The corresponding $O^{i
}_{\text{relevant},j}$ and $O^{i}_{\text{target},j}$ of the correct options along with approximate positions of correct options are used to generate the message \( \zeta^{i} \) using the template: ``I see \{relevant object\} that may be relevant to
your target \{true target\}, and \{possible target object\} may be your target at \{position\}.'' If neither object is identified, the robot will not send a message.

\subsection{Exploration with Communication}\label{sec:baseline_method}

To guide exploration, we first compute the semantic values (SV) $\text{SV}^i_{\text{no-com},j}$ for a 2D grid map as per \cite{ren2024explore}, ignoring the messages received from other agents. 
We denote $P$ as a set of frontier points identified by the VLM at the current pose— locations on the boundary of the explored and unexplored regions \cite{yamauchi1997frontier}.
The SV of ${\hat{p}} \in P$ is denoted as $\text{SV}^i_{{\hat{p}},j}$. 
Upon receiving a message \( \zeta^{\hat{i
}} \),  the SV for each ${\hat{p}} \in P$ based on communication is adjusted based on the counts of the relevant and target objects, scaled by temperature \( \tau_{1} \) and \( \tau_{2} \):
\vspace{-0.15cm}
\begin{equation}
\text{SV}^i_{\text{com},{\hat{p}},j} = \log\left(\tau_{1} \text{Num}(O^i_{\text{relevant},j}) + \tau_{2} \text{Num}(O^i_{\text{target},j})\right),
\end{equation}
where $\tau_1$ and $\tau_2$ weight the counts of relevant and target objects, respectively, balancing indirect semantic cues and direct task information in the communicated semantic value. For each task, the SV is defined as:
\vspace{-0.15cm}
\begin{equation}
\text{SV}^i_{{\hat{p}},j}= \max(\text{SV}^i_{\text{no-com},{\hat{p}},j}, \text{SV}^i_{\text{com},{\hat{p}},j})
\vspace{-0.15cm}
\end{equation}
The semantic value at position $\hat{p}$ is set to the maximum of the agent’s own estimate and the value from received messages, favoring the most informative source. The average of all $\text{SV}^i_{j}$ values is computed to determine the final semantic value:
\vspace{-0.15cm}
\begin{equation}
\text{SV}^i_{\text{final},{\hat{p}}} = 
\frac{1}{N_q} \sum_{j=1}^{N_q} \text{SV}^i_{{\hat{p}},j}
\vspace{-0.15cm}
\end{equation}

If $\text{SV}^i_{\text{final},{\hat{p}}}=0$, the robot randomly selects one of the frontier points to continue exploration. After moving, the robot applies Volumetric Truncated Signed Distance Function (TSDF) Fusion \cite{newcombe2011kinectfusion, zeng20173dmatch} to update the occupancy of the voxels in $M$, which marks them as explored or unexplored based on depth images $I^i_{d}$. This data is projected onto a 2D grid map matching the size of SV maps. The robot then uses Frontier-Based Exploration (FBE) \cite{yamauchi1997frontier} for path planning, where higher SV points indicate the surrounding areas likely to be more informative.
To improve exploration efficiency, Gaussian smoothing spreads each SV point's value across its surrounding region, enabling smoother navigation and helping robots prioritize areas with high information value.

\subsection{Confidence Check}

The robot performs a confidence check to determine if it has sufficient certainty to answer an embodied question. Here, we prompt the VLM to generate answer prediction probabilities over the four possible answers from $\{\text{`A', `B', `C', `D'}\}$, along with a question-image relevance score.  This relevance score reflects the likelihood that the robot’s current view contains the information needed to answer the question.
The VLM's probability of predicting ``Yes" to the prompt ``Consider the {question}. Are you confident about answering the question given the current view?" is regarded as the question-image relevance score 
$\text{Rel}^i_{j}$ for question $q^i_j$.
The set of answer probabilities is defined as $\{\text{Ans}^i_{j}(L) \mid \ L \in \{\text{`A', `B', `C', `D'}\}\}$.
Both $\text{Rel}^i_{j}$ and $\text{Ans}^i_{j}(L)$ are bounded within the interval $[0,1]$. The answer is considered confident, and subsequent exploration will exclude question $q^i_j$, if the following condition is met:
\(
\exists ! L \in \{\text{`A', `B', `C', `D'}\} \ \text{s.t.} \ \text{Ans}^i_{j}(L) \times \text{Rel}^i_{j} > 1 - \epsilon_2,
\)
where \( \epsilon_2 \) is a user-defined threshold that reflects the confidence level needed for termination. This threshold \( \epsilon_2 \) is applied to all robots when determining if an answer can be accepted.

\subsection{Stop Criterion}
In our setup, the robot terminates its exploration once it has answered all the questions assigned to it, either through its own analysis based on its observations and reasoning or with answers provided by the other robots. Alternatively, exploration stops when the maximum allowed time is reached.

\begin{figure*}[!tbp]
    \centering
    \includegraphics[width=\linewidth]{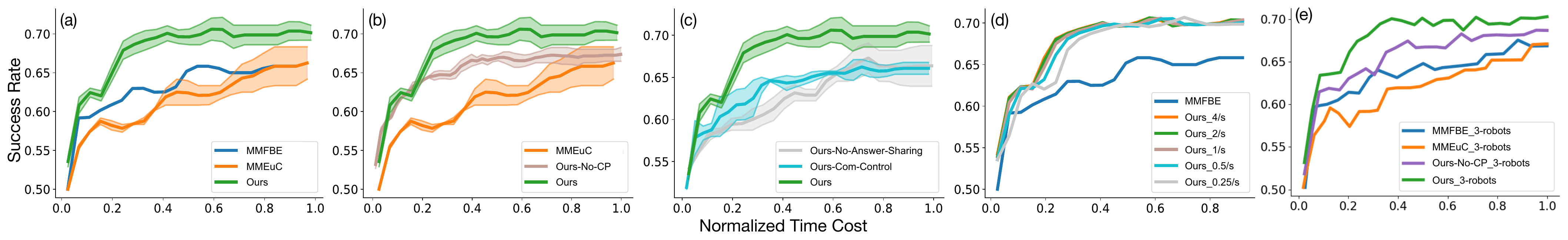}
    \vspace{-0.7cm}
    \caption{The diagrams of SC vs. NTC on our MM-EQA dataset.  (a)–(d) show the results for a 2-robot team. 
    (a) The comparison between our method and baselines. 
    (b) The ablative comparison of communication and conformal prediction modules.
    (c) The ablative comparison of the number of objects in the communication messages and the answering sharing mechanism. 
    (d) The ablative comparison of baselines with our method at message-sending speeds of 0.25, 0.5, 1, 2, and 4 messages per second.  
    (e) A scalability analysis comparing our method and baselines using a 3-robot team.}
    \vspace{-0.6cm}
    \label{fig:result_include1}
\end{figure*}

%% file: contents/05experiments.tex
\subsection{MM-EQA Benchmark}

We create an MM-EQA benchmark based on the HM3D dataset \cite{ramakrishnan2021habitat} to evaluate our framework and baseline methods, which provides photo-realistic, diverse 3D indoor scenarios. 
The corresponding questions are challenging and require semantic reasoning. 
Different from the existing EQA datasets with a single question per scene, we generate six questions for each scene and their corresponding answers to support the multi-agent multi-task setting. 
The embodied questions can be categorized into the following five types:
\begin{enumerate}
    \item \textbf{Location:} This type asks about the location of an object, e.g., ``Where have I left the cushion? A) At the corner of the bedroom B) In the hallway C) Near the basketboard D) Next to TV in the living room''.
    \item \textbf{Identification:} This type asks about identifying the property of an object, e.g., ``What bath mat is in the bathroom? A) Red B) White C) Black D) Gray".
    \item \textbf{Counting:} This type asks about the number of objects, e.g., ``Did I leave any cues or balls on the pool table? A) None B) One C) Two D) Three''.
    \item \textbf{Existence:} This type asks if an object is present at a certain location, e.g., ``Have I put utensils and napkins on the dining table? A) Yes B) No". 
    \item \textbf{State:} This type asks about the state of an object, e.g., ``Is the washing machine turned on? A) Yes B) No". 
\end{enumerate}

We use GPT-4V to generate the task questions and corresponding target objects, with human oversight refining them. Our benchmark includes 420 embodied question tasks across 70 scenarios for two robots, each assigned three tasks for testing. We allocate an additional 20 scenarios to create the calibration dataset for conformal prediction. We use Habitat \cite{puig2023habitat} as the base simulator, which loads 3D scenes.

\subsection{Evaluation Metrics and Baselines}
We evaluate the performance of our method using two metrics: 
\textbf{Success Rate (SR)}, which measures the proportion of correct answers across all questions assigned to the robots, and \textbf{Normalized Time Cost (NTC)}, which quantifies the normalized time taken from the start of navigation to the completion of tasks by all the robots.
The time cost consists of two components: robot movement and message sending. 
To analyze the impact of communication latency, we experiment with several different message-sending speeds. By default, we set the robot's movement speed to \SI{1}{m/s} and the message-sending speed to one per second to maintain consistency and comparability across experiments.

We compare our method against the following baselines and ablation settings:
\begin{enumerate}
    \item \textbf{MMFBE:} A multi-agent multi-task frontier-based exploration method extended from the canonical FBE \cite{yamauchi1997frontier}. MMFBE employs a VLM to answer questions but does not use it for semantic mapping during exploration and does not involve communication.
    \item \textbf{MMEuC:} A multi-robot multi-task extension of the Explore Until Confident (EUC) framework \cite{ren2024explore}, where robots operate independently without communication.
    \item \textbf{Ours-No-CP:} An ablation that allows communication but omits conformal prediction, enabling evaluation of its contribution to calibrating LLM confidence.
    \item \textbf{Ours-Com-Control:} An ablation that controls the number of objects included in communication messages. To match the effect of CP, we fix the number of objects to the same level as with CP and randomly sample from observed objects.
    \item \textbf{Ours-No-Answer-Sharing:} An ablation where robots exchange observations and calibrated predictions relevant to each other's tasks but are not sharing answers, isolating the impact of answer-sharing on performance.
\end{enumerate}

\begin{figure*}[!tbp]
	\centering
        \includegraphics[width=0.95\linewidth]{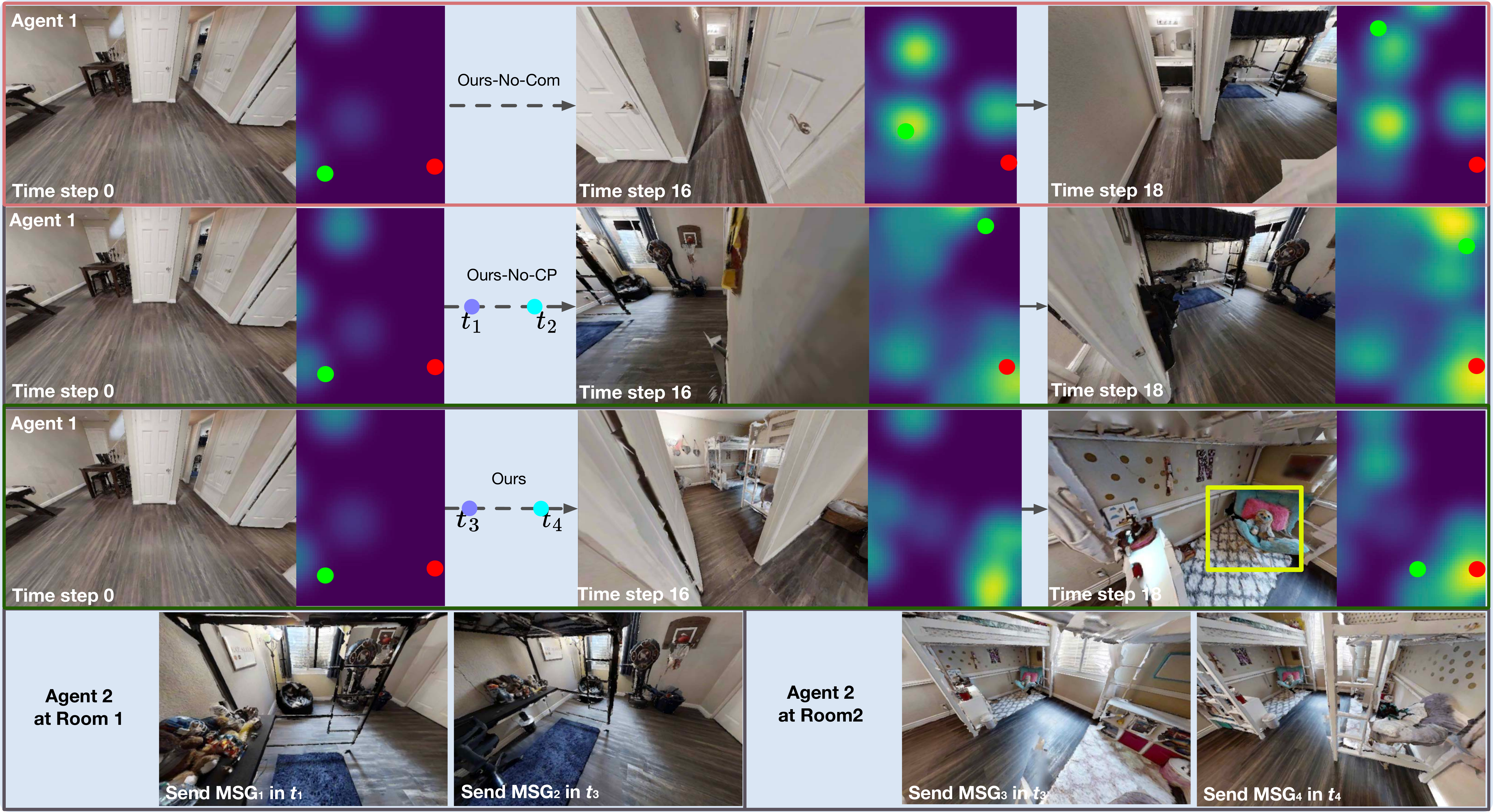}
        \vspace{-0.2cm}
	\caption{The comparisons of robot views and global SV maps among three methods. The red points represent the location of the target and the green points represent the position of the robot. Agents in the three methods start from the same pose. The question for this scenario is ``Where is the red bear cushion?" For ``Ours-No-CP" and ``Ours", Robot2 separately explores the same rooms at different times and sends messages to Robot1.
    The detailed messages are as follows: 
    $\text{MSG}_{1}$: I see a basketboard, dolls, black chair that may be relevant to your target red bear cushion, and dolls may be your target at $\{position 1\}$. $\text{MSG}_{2}$: I see dolls that may be relevant to your target red bear cushion at $\{position 2\}$. $\text{MSG}_{3}$: I see bed, red pillow on blue chair that may be relevant to your target red bear cushion, and red pillow on blue chair may be your target at $\{position 3\}$. $\text{MSG}_{4}$: I see red pillow on blue chair that may be relevant to your target red bear cushion, and a red pillow on blue chair may be your target at $\{position 4\}$. 
}
\vspace{-0.4cm}
	\label{fig:realview}
\end{figure*}

\begin{figure}[!tbp]
    \centering
    \includegraphics[width=0.85\linewidth]{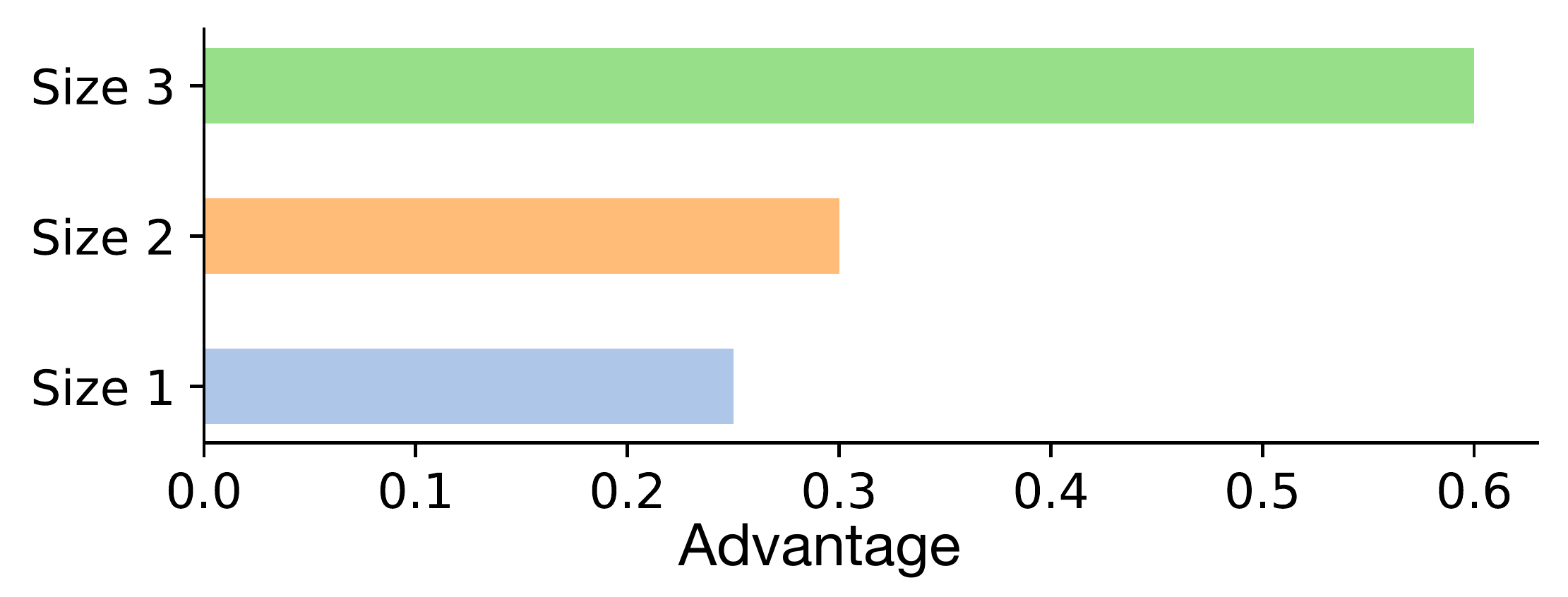}
    \vspace{-0.3cm}
    \caption{The comparison of performance improvement in the environments with different sizes. The ``Advantage" represents the difference between the NTC of ``Ours" and the NTC of MMFBE, calculated as \( \text{Advantage} = \text{NTC}_{\text{Ours}} - \text{NTC}_{\text{MMFBE}} \). Size 1 represents scene area $L \times W < 150 \, \text{m}^2$. Size 2 represents $150 \leq L \times W < 250 \, \text{m}^2$. Size 3 represents $L \times W \geq 250 \, \text{m}^2$.}
    \vspace{-0.5cm}
    \label{fig:location}
\end{figure}
\vspace{-0.1cm}
\subsection{Implementation Details}
We use the same VLM (Prismatic-VLM-13B) \cite{karamcheti2024prismatic} as in \cite{ren2024explore}. Since our setting requires probability outputs from the LLM, we cannot use state-of-the-art closed-source models like GPT-4V. Instead, we employ LLaMA3-8B-instruct, a smaller open-source model. We set the temperature of LLM to 0.7. We set the $\tau_{1}, \tau_{2}$ to be 1.0 and 10.0, respectively. The non-conformity threshold $p_\text{thres}$ is computed as the $(1 - \epsilon_1)$ quantile of the calibration set probabilities, where $\epsilon_1$ controls the prediction set size. In our setup, this corresponds to a 0.6 quantile for Option A and a 0.82 quantile for Option B.
We conduct all experiments using two NVIDIA 6000 Ada Generation GPUs, with VLM and LLM on separate devices.

\subsection{Results and Analysis}

\textbf{Communication Effectiveness.} 
We demonstrate the effectiveness of our communication module through the average success rate achieved when the robots are allowed to explore within different time horizons, which is shown in Fig.~\ref{fig:result_include1}(a). 
Our method significantly outperforms the baseline, MMFBE, by achieving an SR of 0.68 at an NTC of 0.4, compared to MMFBE's SR of 0.65 at an NTC of 0.8, effectively doubling the efficiency. 
This substantial efficiency gain comes from our method's ability to share relevant, calibrated information, allowing agents to prioritize the exploration of important areas. 
In contrast, the ``MMEuC" baseline underperforms even MMFBE because it lacks communication between agents. Without communication, robots are unable to coordinate their efforts or share insights, leading to redundant or inefficient exploration paths.
Additionally, ``Ours" achieves the highest final success rate at convergence, which implies the effectiveness of our communication strategy.
In terms of actual time spent, ``Ours" completes all questions in an average of 445 seconds, compared to 594 seconds for MMFBE, demonstrating faster task completion.

The superior performance of our approach is attributed to the strategic exchange of information, which enables agents to collectively explore relevant objects rather than randomly searching without context. This is a clear advantage over MMFBE which lacks heuristic-based guidance and operates without the benefit of collaborative data sharing. It is also notable that as the NTC increases and more objects are explored, there is greater diversity in the LLM's output, consequently causing an increase in the standard deviation of the results. There is no deviation in MMFBE since FBE is a deterministic rule-based method.

In Fig.~\ref{fig:result_include1}(c), the comparison with the ``Ours-No-Answer-Sharing" baseline shows the impact of sharing answers to other robots' questions. Without this ability, a robot spends additional time exploring answers to questions that could be addressed by its partners. This lack of efficiency is reflected in the higher NTC and a lower SC compared to the full ``Ours" configuration. Consequently, the absence of this capability results in slower convergence and reduced overall task efficiency, as each agent must independently answer questions without fully leveraging high-confidence information shared by its partners. By enabling answer sharing, our method ensures that high-confidence shared information is leveraged to facilitate task completion.

\textbf{Confidence Calibration with Conformal Prediction.} 
We illustrate the importance and effectiveness of CP in Fig.~\ref{fig:result_include1}(b). 
Without CP, the ``Ours-No-CP" baseline achieves performance similar to ``MMEuC", showing that uncalibrated outputs mislead agents with ineffective or erroneous information, leading to exploration in less relevant areas and reduced efficiency. In contrast, the calibrated communication in ``Ours" ensures that messages are reliable, improving task success.
Fig.~\ref{fig:result_include1}(c) shows that the ``Ours-Com-Control" configuration, which sends more frequent but less relevant information, fails to improve efficiency, implying that the \textit{quality} of shared information is more critical than its \textit{quantity}. Ineffective communication, providing more information with less relevance in ``Ours-Com-Control", results in worse performance and slower convergence, with the final SR similar to those in the ``MMEuC'' configuration.

We demonstrate a representative test case in Fig.~\ref{fig:realview}. In the ``MMEuC" configuration, the SV map modes diffuse slowly and do not cover the important areas, which leads to a low probability of the robots finding target objects. This is because they tend to move to irrelevant areas rather than exploring the correct ones, decreasing the chances of finding targets. Although the modes of the SV map in ``Ours-No-CP" diffuse more rapidly than in ``Ours", the uncalibrated communication causes the robot to navigate to the wrong room due to misleading information from miscalibrated messages. In contrast, ``Ours" effectively updates the ego robot's semantic value map, guiding the robot efficiently and demonstrating the importance of calibrated communication.

\textbf{Impact of Scene Size.} Fig.~\ref{fig:location} further highlights how scene size affects the performance of ``Ours" compared to the MMFBE baselines. As the scene area increases, the advantage of our communication method becomes more pronounced. In larger scenes (Size 3), our method achieves an average NTC improvement of 0.6 over MMFBE, showcasing a substantial efficiency gain due to enhanced information sharing and coordinated exploration. 
The MMFBE baseline struggles more as scene size increases because it relies on rule-based, non-communicative exploration, which lacks adaptability to larger, more complex environments. 
Our method, leveraging calibrated and relevant shared information, consistently enhances exploration efficiency and task success, which demonstrates its robustness and scalability.

\textbf{Impact of Communication Latency.} 
We evaluate task success under different message-sending speeds to analyze the effect of communication latency. Fig.~\ref{fig:result_include1}(d) shows that higher sending speeds lead to faster increases in success rates in early stages by enabling faster information exchange. After sufficient exploration, the final success rates under different speeds become almost identical. Notably, our approach outperforms the MMFBE baseline across all message-sending speeds, demonstrating its effectiveness regardless of latency.

\textbf{Scalability Analysis.}
We evaluate scalability by extending to a three-robot team. As shown in Fig.~\ref{fig:result_include1}(e), ``Ours'' achieves a faster increase in SR with respect to NTC compared to the baselines. Conversely, ``Ours-No-CP'' suffers a decline in early-stage SR due to the increased presence of irrelevant information. While all methods benefit from more agents over time, our method consistently completes tasks more efficiently and scales with minimal computational overhead.

%% file: contents/06conclusions.tex
To improve the communication efficiency of decentralized LLM-based multi-agent systems, we propose an effective natural language communication framework with LLMs. Due to the hallucinations of LLMs and their potentially negative effects on overall efficiency, our framework incorporates conformal prediction to calibrate the confidence of outputs, which filters out overconfident LLM-generated outputs and makes the information in communication effective.
The experimental results demonstrate that our method significantly improves exploration efficiency and the overall success rate, particularly in larger scenes.  In this work, we develop our method in both two- and three-agent cooperative settings through efficient communication. Future work will involve scaling our method to more complex deployments with significantly larger robot teams and environments.